\documentclass[11pt,a4paper]{article}
\usepackage[hyperref]{naaclhlt2018}
\usepackage{times}
\usepackage{latexsym}
\usepackage{verbatim}
\usepackage[inline, shortlabels]{enumitem}
\usepackage{amsmath,amsfonts,amssymb} 
\usepackage{arydshln}
\usepackage{color}
\usepackage{subcaption}
\usepackage{multirow}
\usepackage{graphicx}
\usepackage{enumitem}
\usepackage{multicol}
\usepackage{url}
\usepackage{threeparttable, tablefootnote}
\usepackage[hang,flushmargin]{footmisc}
\usepackage[export]{adjustbox}
\usepackage{algorithm}
\usepackage{algpseudocode}

\DeclareCaptionFont{10pt}{\fontsize{10pt}{12pt}\selectfont}
\aclfinalcopy 
\def\aclpaperid{61} 

\renewcommand{\footnotesize}{\fontsize{8pt}{11pt}\selectfont}

\title{Self-Attentive Residual Decoder for Neural Machine Translation}

\author{Lesly Miculicich Werlen\textsuperscript{*,$\dagger$}, Nikolaos Pappas\textsuperscript{*},   Dhananjay Ram\textsuperscript{*,$\dagger$},\\ \textbf{Andrei Popescu-Belis\textsuperscript{$\ddagger$}} \\
\textsuperscript{*}{Idiap Research Institute, Switzerland}, \\
\textsuperscript{$\dagger$}{\'{E}cole polytechnique f\'{e}d\'{e}rale de Lausanne (EPFL), Switzerland}, \\
\textsuperscript{$\ddagger$}{HEIG-VD / HES-SO, Switzerland}\\
{\tt \{lmiculicich, npappas, dram\}@idiap.ch} \\
{\tt andrei.popescu-belis@heig-vd.ch}}

\date{}

\begin{document}
\maketitle

\begin{abstract}
Neural sequence-to-sequence networks with attention have achieved remarkable performance for machine translation. One of the reasons for their effectiveness is their ability to capture relevant source-side contextual information at each time-step prediction through an attention mechanism.
However, the target-side context is solely based on the sequence model which, in practice, is prone to a recency bias and lacks the ability to capture effectively non-sequential dependencies among words.
To address this limitation, we propose a target-side-attentive residual recurrent network for decoding, where attention over previous words contributes directly to the prediction of the next word. The residual learning facilitates  the flow of information from the distant past and is able to emphasize any of the previously translated words, hence it gains access to a wider context.
The proposed model outperforms a neural MT baseline as well as a memory and self-attention network on three language pairs. 
The analysis of the attention learned by the decoder confirms that it emphasizes a wider context, and that it captures syntactic-like structures.
\end{abstract}
\vspace{1mm}
\section{Introduction}
Neural machine translation (NMT) has recently become the state-of-the-art approach to machine translation \cite{bojar-EtAl:2016:WMT1}.
Several architectures have been proposed for this task \cite{kalchbrenner13, Sutskever14, cho-EtAl:2014:EMNLP2014, gehring2017convolutional, vaswani2017attention}, but the attention-based NMT model 
 designed by \citet{bahdanau2014neural} is still considered the de-facto baseline. This architecture is composed of two recurrent neural networks (RNNs), an encoder and a decoder, and an attention mechanism between them for modeling a soft word-alignment. 
First, the model encodes the complete source sentence, and then decodes one word at a time. The decoder has access to all the context on the source side through the attention mechanism.
However, on the target side, the contextual information is represented only through a fixed-length vector, namely the hidden state of the decoder.  
As observed by \citet{bahdanau2014neural}, this creates a bottleneck which hinders the ability of the sequential model to learn longer-term information effectively. 

\begin{figure}[t]
	\centering
	\begin{subfigure}[b]{0.49\linewidth}
		\centering
		\includegraphics[width=1\linewidth]{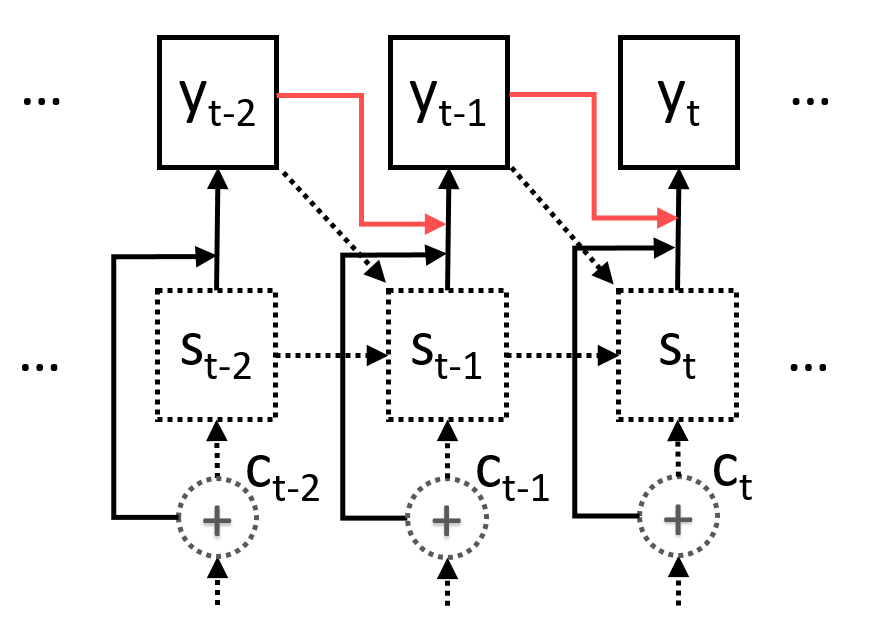}
		\caption{Baseline NMT decoder} 
		\label{fig:baseline}
	\end{subfigure}
	\begin{subfigure}[b]{0.49\linewidth}
		\centering
		\includegraphics[width=1\linewidth]{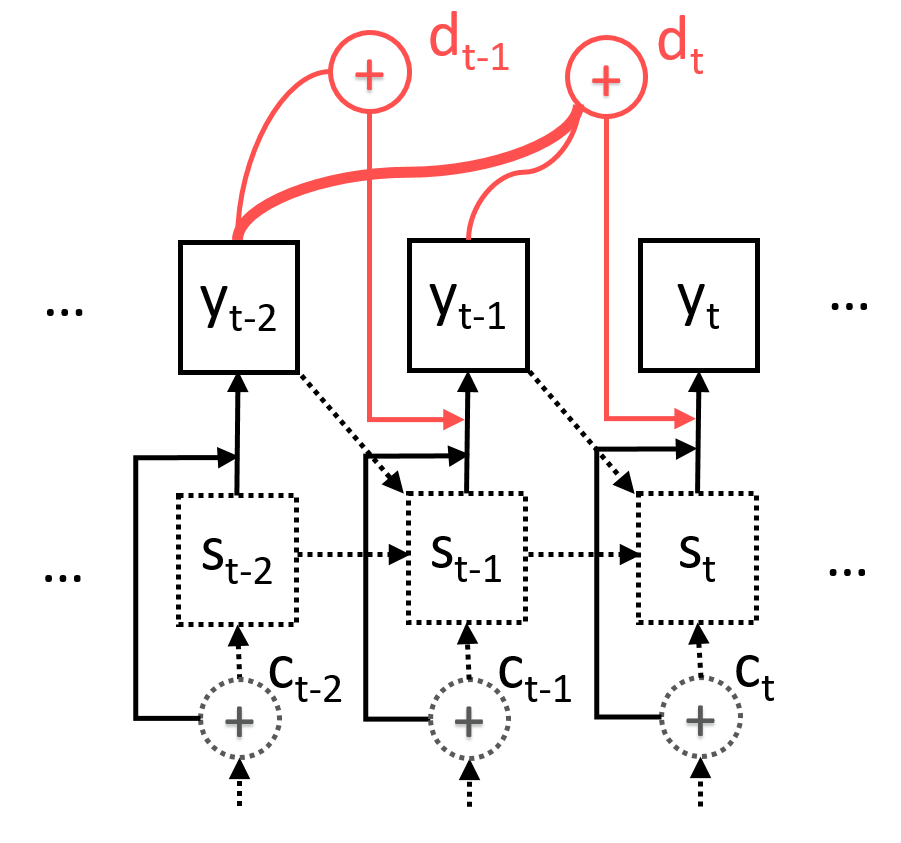}
		\caption{Self-attentive residual dec.}
		\label{fig:attention}
	\end{subfigure}
	\caption{Comparison between the decoder of the baseline NMT and the proposed decoder with self-attentive residual connections.}
\end{figure}

As pointed out by \citet{cheng-dong-lapata:2016:EMNLP2016}, sequential models present two main problems for natural language processing. First,
the memory of the encoder is shared across multiple words and is prone to bias towards the recent past. Second, such models do not fully capture the structural composition of language.
To address these limitations, several recent models have been proposed, namely memory networks \cite{cheng-dong-lapata:2016:EMNLP2016,tran-bisazza-monz:2016:N16-1,wang-EtAl:2016:EMNLP20161} and self-attention networks \cite{daniluk2016frustratingly, liu2017learning}.  We experimented with these methods, applying them to NMT: \emph{memory RNN} \cite{cheng-dong-lapata:2016:EMNLP2016} and \emph{self-attentive RNN}  \cite{daniluk2016frustratingly}. However, we observed no significant gains in performance over the baseline architecture.  

In this paper, we propose a self-attentive residual recurrent decoder, presented in Figure~\ref{fig:attention}, which, if unfolded over time, represents a densely-connected residual network.  The self-attentive residual connections focus selectively on previously translated words and propagate useful information to the output of the decoder, within an attention-based NMT architecture. The attention paid to the previously predicted words is analogous to a read-only memory operation, and enables the learning of syntactic-like structures which are useful for the  translation task. 

Our evaluation on three language pairs shows that the proposed model improves over several baselines, with only a small increase in computational overhead. In contrast, other similar approaches 
have lower scores but a
higher computational overhead. The  contributions of this paper can be summarized as follows:
\begin{itemize}[noitemsep, leftmargin=4mm]
 \item We propose and compare several options for using self-attentive residual learning within a standard decoder, which facilitates the flow of contextual information on the target side.
 \item We demonstrate consistent improvements over a standard  baseline, and two advanced variants, which make use of memory and self-attention on three language pairs (English-to-Chinese, Spanish-to-English, and English-to-German).
 \item We perform an ablation study and analyze the learned attention function, providing additional insights on its actual contributions.
\end{itemize}


\section{Related Work}\label{sec:related}

Several studies have been proposed to enhance sequential models by capturing longer contexts. Long short-term memory (LSTM) \cite{hochreiter1997long} is the most commonly used recurrent neural network (RNN), because its internal memory allows to retain information from a more distant past than a vanilla RNN. 
Several studies attempt to increase the memory capacity of LSTMs by using memory networks \cite{weston2014memory, sukhbaatar2015end}.  For instance, \citet{cheng-dong-lapata:2016:EMNLP2016} incorporate different memory cells for each previous output representation, which are later accessed by an attention mechanism. \citet{tran-bisazza-monz:2016:N16-1} include a memory block to access recent input words in a selective manner. Both methods show improvements on language modeling. For NMT, \citet{wang-EtAl:2016:EMNLP20161} presented a decoder enhanced with an external shared memory. Memory networks extend the capacity of the network and have the potential to read, write, and forget information. Our method, which attends over previously predicted words, can be seen as a read-only memory, which is simpler but computationally more efficient because it does not require additional memory space. 

Other studies aim to improve the modeling of
source-side contextual information, for example 
through a context-aware encoder using self-attention \cite{zhang2017context}, 
or a recurrent attention NMT \cite{yang-EtAl:2017:EACLshort1} that is
aware of previously attended words on the source-side in order to better predict which words will be attended in future.
Additionally,
variational NMT \cite{zhang-EtAl:2016:EMNLP20162} introduces a latent variable to model the underlying semantics of source sentences. In contrast to these studies, we focus instead on the contextual information \emph{on the target side}. 

The application of self-attention mechanisms to RNNs have been previously studied, and in general, they seem to capture syntactic dependencies among distant words \cite{liu2017learning, soltani2016higher, lee2017recurrent, lin2017structured}. \citet{daniluk2016frustratingly} explore different approaches to self-attention for language modeling, leading to improvements over a baseline LSTM and over memory-augmented methods. However, the methods do not fully utilize a longer context. The main difference with our approach is that we apply attention on the output embeddings rather than the hidden states.  Thus, the connections are independent of the recurrent layer representations, which is beneficial to NMT, as we show below. 

Our model relies on residual connections, which have been shown to improve the learning process of deep neural networks by addressing the vanishing gradient problem \cite{he2016deep}. These connections create a direct path from previous layers, helping the transmission of information. Recently, several architectures using residual connections with LSTMs have been proposed for sequence prediction \cite{zhang2016highway,kim2017residual, zilly2016recurrent, wang-tian:2016:EMNLP2016}. To our knowledge, our study is the first one to use self-attentive residual connections within residual RNNs
for NMT. In parallel to our study, a similar method was recently proposed for sentiment analysis \cite{wang2017rra}.

\section{Background: Neural Machine Translation}\label{sec:base}
Neural machine translation aims to compute the conditional distribution of emitting a sentence in a target language given a sentence in a source language, denoted by \(p_{\Theta}( y | x)\), where \(\Theta\) is the set of parameters of the neural model, and \(y =\{y_1,...,y_n\}\) and \(x =\{x_1,...,x_m\}\) are respectively the representations of source and target sentences as sequences of words.  The parameters $\Theta$ are learned by training a sequence-to-sequence neural model on a corpus of parallel sentences. In particular, the learning objective is to maximize the following conditional log-likelihood:
\begin{equation}
 \max_{\Theta} \frac{1}{N} \sum_{n=1}^N \log(p_{\Theta}(y|x))
\end{equation}
The models typically use gated recurrent units (GRUs) \cite{cho-EtAl:2014:EMNLP2014} or LSTMs \cite{hochreiter1997long}.  Their architecture has three main components: an encoder, a decoder, and an attention mechanism.

The goal of the encoder is to build meaningful representations of the source sentences. It consists of a bidirectional RNN which includes contextual information from past and future words into the vector representation $h_i$ of a particular word vector $x_i$, formally defined as follows:
\begin{align}
 h_i &=[\overrightarrow{h_i}, \overleftarrow{h_i}]
\end{align}
Here, \(\overrightarrow{h_i}= f(x_i, h_{i-1})\) and \(\overleftarrow{h_i}= f(x_i, h_{i+1})\) are the hidden states of the forward and backward passes of the bidirectional RNN respectively, and \(f\) is a non-linear function.

The decoder (see Figure~\ref{fig:baseline}) is in essence a recurrent language model. At each time step, it predicts a target word \(y_t\) conditioned over the previous words and the information from the encoder using the following posterior probability:
\begin{equation}\label{dec_prob}
p(y_t|y_1,...,y_{t-1}, c_t) \approx g(s_t, y_{t-1}, c_t)
\end{equation}
where \(g\) is a non-linear multilayer function.
The hidden state of the decoder \(s_t\) is defined as:
\begin{equation}\label{dec_hid}
s_t = f(s_{t-1}, y_{t-1}, c_t)
\end{equation}
and depends on a \emph{context vector} \(c_t\) that is computed by the attention mechanism.

The attention mechanism allows the decoder  to select which parts of the source sentence are more useful to predict the next output word. This goal is achieved by considering a weighted sum over all hidden states of the encoder as follows:
\begin{gather}
	c_t=\sum_{i=1}^{m} \alpha^t_i h_i
\end{gather}
where \(\alpha^t_{i}\) is a weight calculated using a normalized exponential function $a$, also known as \emph{alignment function}, which computes how good is the match between the input at position \(i \in \{1,...,n\}\) and the output at position \(t\):
\begin{gather}
  \alpha^t_i =  softmax(e_i^t) \\
  e_i^t = a(s_{t-1}, h_i)
\end{gather}
Different types of alignment functions have been used for 
NMT, as investigated by \citet{D15-1166}. Here, we use the one originally defined by \citet{bahdanau2014neural}.

\section{Self-Attentive Residual Decoder}\label{sec:dec}
The decoder of the attention-based NMT model uses a skip connection from the previously predicted word to the output classifier in order to enhance the performance of translation. As we can see in Eq.~\eqref{dec_prob}, the probability of a particular word is calculated by a function $g$ which takes as input the hidden state of the recurrent layer $s_t$, the representation of the previously predicted word $y_{t-1}$, and the context vector $c_t$.
Within $g$, these quantities are typically summed up after going through simple linear transformations, hence the addition of $y_{t-1}$ is indeed a skip connection as in residual networks \cite{he2016deep}. In theory, $s_t$ should be sufficient for predicting the next word given that it is dependent on the other two local-context components according to Eq.~\eqref{dec_hid}.
However, the $y_{t-1}$ quantity makes the model emphasize the last predicted word for generating the next word.
How can we make the model consider a broader context?

To answer this question, we propose to include into the decoder's formula skip connections not only from the previous time step $y_{t-1}$, but from all previous time steps from $y_0$ to $y_{t-1}$.  This defines a residual recurrent network which, unfolded over time,  can be seen as a densely connected residual network. These connections are applied to all previously predicted words, and reinforce the memory of the recurrent layer towards what has been translated so far. At each time step, the model decides which of the previously predicted words should be emphasized to predict the next one.  In order to deal with the dynamic length of this new input, we use a target-side summary vector $d_t$  that can be interpreted as the representation of the decoded sentence until the time $t$ in the word embedding space.  We therefore modify Eq.~\eqref{dec_prob} replacing $y_{t-1}$ with $d_t$:
\begin{equation}
p(y_t|y_1,...,y_{t-1}, c_t) \approx g(s_t, d_{t},  c_t)
\end{equation}
The replacement of $y_{t-1}$ with $d_t$ means that
the number of parameters added to the model is dependent only on the calculation of $d_t$. Figure~\ref{fig:attention} illustrates the change made to the decoder.
We define two methods for summarizing the context into $d_t$, which are described in the following sections.

\subsection{Mean Residual Connections}
One simple way to aggregate information from multiple word embeddings is by averaging them.
This average can be seen as the sentence representation until time $t$.
We hypothesize that this representation is more informative than using only the embedding of the previous word. Formally:
\begin{equation}\label{eq:d-avg}
d_t^{avg} = \frac{1}{t-1} \sum_{i=1}^{t-1} y_i
\end{equation}

\subsection{Self-Attentive Residual Connections}\label{sec:attrc}
Averaging is a simple and cheap way to aggregate information from multiple words, but may not be sufficient for all kinds of dependencies.  Instead, we propose a dynamic way to aggregate information in each sentence, such that different words have different importance according to their relation with the prediction of the next word.  We propose to use a shared self-attention mechanism to obtain a summary representation of the translation, i.e.\ a \emph{weighted average representation} 
of the words translated  from $y_0$ to $y_{t-1}$.
This mechanism aims to model, in part, important non-sequential dependencies among words, and serves as a complementary memory to the recurrent layer.
\begin{gather}\label{eq:d-cavg}
d_t^{cavg} = \sum_{i=1}^{t-1} \alpha_i^t y_i \\
 \alpha_i^t =  softmax(e_i^t)
\end{gather}
The weights of the attention model are computed by a scoring function \(e_i^t\) that predicts how important each previous word ($y_0, ...,$ or $y_{t-1}$) is for the current prediction $y_t$. 

We experiment with two different scoring functions, as follows:
 \begin{equation}\label{eq:scoring-functions} 
  e_i^t = v^\intercal tanh(W_y y_i + W_s s_t)\ \ \text{\small \emph{(content+scope)}}
\end{equation} \vspace{-4mm}
\begin{equation}
  \mathrm{or}\ \ e_i^t = v^\intercal tanh(W_y y_i)\ \ \text{\small \emph{(content)}}
 \end{equation}
where $v \in \mathbb{R}^{e}$, $W_y \in \mathbb{R}^{e \times e}$, and $W_s \in \mathbb{R}^{e \times d}$ are weight matrices, $e$ and $d$ are the dimensions of the embeddings and hidden states respectively.
Firstly, we study the scoring function noted \emph{content+scope}, as proposed by \citet{bahdanau2014neural} for NMT. Secondly, we explore a scoring function noted as \emph{content}, which is calculated based only on the previous hidden states of the decoder, as proposed by \citet{pappas2017explicit}.  In contrast to the first attention function, which makes use of the hidden vector $s_t$, the second one is based only on the previous word representations, therefore, it is independent of the current prediction representation. However, the normalization of this function still depends on $t$.

\section{Other Self-Attentive Networks}\label{sec:other} 
To compare our approach with similar studies, we adapted two representative self-attentive networks for application to NMT.

\subsection{Memory RNN}
The \emph{Memory RNN} decoder is based on the proposal by \citet{cheng-dong-lapata:2016:EMNLP2016} to modify an LSTM  
 layer to include a memory with different cells for each previous output representation. Thus at each time step, the hidden layer can select past information dynamically from the memory. 
 To adapt it to our framework, we modify Eq.~\eqref{dec_hid} as:
\begin{gather}
s_t = f(\tilde{s}_t, y_{t-1}, c_t)
\end{gather}
\begin{flalign}
\text{where} \quad \quad   &\tilde{s}_t =\sum_{i=1}^{t-1} \alpha^t_i s_i \\
	&\alpha^t_i = softmax(e_i^t) \\
	&e_i^t = a(h_i, y_{t-1},  \tilde{s}_{t-1})
\end{flalign}

\subsection{Self-Attentive RNN}
The \emph{Self-Attentive RNN} is the simplest one proposed by \citet{daniluk2016frustratingly}, and incorporates a summary vector from past predictions calculated with an attention mechanism. Here, the attention is applied over previous hidden states. This decoder is formulated as follows:
\begin{equation}
 p(y_t|y_1,...,y_{t-1}, c_t)  \approx g(s_t, y_{t-1},  c_t, \tilde{s}_t)
\end{equation}
\begin{align}
\text{where} \quad \quad  \tilde{s}_t &= \sum_{i=1}^{t-1} \alpha_i^t s_i \\
\alpha_i^t  &=  softmax(e_i^t)\\
e_i^t  &= a(s_i, s_{t})
\end{align}

Additional details of the formulations in Sections~\ref{sec:base}, \ref{sec:dec}, and \ref{sec:other} are described in the Appendix~\ref{app}.

\section{Experimental Settings}

\subsection{Datasets}
To evaluate the proposed MT models in different conditions, we select three language pairs with increasing amounts of training data:  English-Chinese (0.5M~sentence pairs), Spanish-English (2.1M), and English-German~(4.5M).

For English-to-Chinese, we use a subset of the UN parallel corpus \cite{rafalovitch2009united}\footnote{\url{http://www.uncorpora.org/}}, with 0.5M sentence pairs for training, 2K for development, and 2K for testing.  For training Spanish-to-English MT, we use a subset of WMT 2013 \cite{bojar-EtAl:2013:WMT}, corresponding to Europarl v7 and News Commentary v11 with ca.\ 2.1M sentence pairs. Newstest2012 and Newstest2013 were used for development and testing respectively. Finally, we use the complete English-to-German set from WMT 2016 \cite{bojar-EtAl:2016:WMT1} with a total of ca.\ 4.5M sentence pairs. The development set is Newstest2013, and the testing set is Newstest2014. Additionally, we include as testing sets Newstest2015 and Newstest2016, for comparison with the state of the art.
We report translation quality using (a)~BLEU over \emph{tokenized} and \emph{truecased} texts, and (b)~NIST BLEU over \emph{detokenized} and \emph{detruecased} texts\footnote{Scrips from Moses toolkit \cite{koehn-EtAl:2007:PosterDemo}: BLEU \emph{multi-bleu}, NIST BLEU \emph{mteval-v13a.pl}, \emph{tokenizer.perl}, \emph{truecase.perl}.}. 

\subsection{Model Configuration}
We use the implementation of the attention-based NMT baseline provided in \texttt{team2016theano}\footnote{\url{https://github.com/nyu-dl/dl4mt-tutorial}} developed in Python using Theano \cite{2016arXiv160502688short}.  The system implements an attention-based NMT model, described above, using one layer of GRUs \cite{cho-EtAl:2014:EMNLP2014}. The vocabulary size is 25K for English-to-Chinese NMT, and 50K for Spanish-to-English and English-German.  We use the byte pair encoding (BPE) strategy for out-of-vocabulary words \cite{sennrich-haddow-birch:2016:P16-12}.  For all cases,  the maximum sentence length of the training samples is 50, the dimension of the word embeddings is 500, and the dimension of the hidden layers is 1,024. We use dropout with a probability of 0.5 after each layer. The parameters of the models are initialized randomly from a standard normal distribution scaled to a factor of 0.01. The loss function is optimized using Adadelta \cite{zeiler2012adadelta} with
$\epsilon=10^{-6}$ and $\rho=0.95$ as in the original paper.
The systems were trained in 7--12 days for each model on a Tesla K40 GPU at the speed of about 
1,000 words/sec.

\section{Analysis of the Results}
Table~\ref{tab:bleu} shows the BLEU scores and the number of parameters used by the different NMT models. Along with the NMT baseline, we included a statistical machine translation (SMT) model based on Moses \cite{koehn-EtAl:2007:PosterDemo} with the same training/tuning/test data as the NMT. The performance of \emph{memory RNN} is similar to the baseline and, as confirmed later, its focus of attention is mainly on the prediction at $t-1$. The \emph{self-attentive RNN} method is inferior to the baseline, which can be attributed to the overhead on the hidden vectors that have to learn the recurrent representations and the attention simultaneously. The proposed models outperform the baseline, and the best scores are obtained by the NMT model with \emph{self-attentive residual connections}. Despite their simplicity, the \emph{mean residual connections} already improve the translation, without increasing the number of parameters.

\begin{table}[t]
	\center
	\small
	{\def\arraystretch{1.2}\tabcolsep=1pt
		\begin{tabular}{l c c c }
			& \textbf{$\pmb{|\Theta|}$} &  \multicolumn{2}{c}{\textbf{BLEU}} \\ \hline
			\textbf{Models} & &\textbf{En\textendash Zh}& \textbf{Es\textendash En}  \\ [1mm] \hline
			SMT baseline & -- & 21.6 & 25.2    \\
			NMT baseline & 108.7M & 22.6 & 25.4  \\ \hdashline
			+ Memory RNN & 109.7M  & 22.5 &  25.5   \\
			+ Self-attentive RNN & 110.2M & 22.0  &  25.1  \\ \hdashline
			+ Mean residual connections &  108.7M  & 23.6 & 25.7   \\
			+ Self-attentive residual connections &  108.9M  & \textbf{24.0} & \textbf{26.3}  \\ \hline
	\end{tabular}}
	\caption{BLEU score (multi-bleu) on \emph{tokenized} text. The highest score per dataset is marked in bold. The self-attentive residual connections make use of the \emph{content} attention function. $|\Theta|$ indicates the number of parameters per model.}
	\label{tab:bleu}
	
\end{table}

\begin{table}[t]
	\center
	\small
	{\def\arraystretch{1.2}\tabcolsep=1.1pt
		\begin{tabular}{l c c}
			&  \multicolumn{2}{c}{\textbf{BLEU }} \\ \hline
			\textbf{Models} &  \textbf{NT14} & \textbf{NT15}  \\ [1mm] \hline			
			NMT (unk. word repl.) \cite{D15-1166} & 20.9 & -- \\				 
			Context-aware NMT \cite{zhang2017context} & 22.57 & -- \\
			Recurrent attention NMT \cite{yang-EtAl:2017:EACLshort1} & 22.1 & 25.0 \\
			Variational NMT \cite{zhang-EtAl:2016:EMNLP20162} & -- & 25.49 \\ \hdashline
			NMT baseline & 22.3 & 24.8 \\
			+ Memory RNN & 22.6 & 24.9\\
			+ Self-attentive RNN &  22.0 & 24.3 \\ \hdashline
			+ Mean residual connections &  22.9 & 24.9  \\
			+ Self-attentive residual connections  & \textbf{23.2} & \textbf{25.5} \\ \hline
	\end{tabular}}
	\caption{BLEU score (multi-bleu) on \emph{tokenized} text for English-to-German on \emph{Newstest (NT) 2014, and 2015}. The highest score per dataset is marked in bold. The self-attentive residual connections makes use of the \emph{content} attention function.}
	\label{tab:bleu3}
	
\end{table}

\begin{table}[t]
	\center
	\small
	{\def\arraystretch{1.2}\tabcolsep=1.5pt
		\begin{tabular}{l c c c c}
			&  \multicolumn{4}{c}{\textbf{BLEU (NIST)}}  \\ \hline
			\textbf{Models} &  \textbf{NT14} &  \textbf{NT15} &  \textbf{NT16}  \\ [1mm] \hline
			Winning WMT &20.1  & 24.4  & \textbf{34.2}  \\ 
			NMT (BPE) \cite{sennrich-haddow-birch:2016:P16-12} & -- & 22.8 & -- \\
			Syntax NMT \cite{nadejde2017predicting}  & -- & -- & 29.0 \\ \hdashline
			NMT Baseline  & 21.0 & 24.4 & 28.8 \\
			+ Mean residual connections*  & 21.4 & 24.7 & 29.6  \\
			+ Self-attentive residual connections**  & \textbf{21.7} & \textbf{25.0} &   29.7 \\ \hline
	\end{tabular}}
	\caption{NIST BLEU scores on \emph{detokenized} and  \emph{detruecased} text for English-to-German on \emph{Newstest (NT) 2014, 2015, 2016}. Significance test: * $p < 0.05$, ** $p<0.01$. The Winning WMT systems are listed in the text below.} 
	\label{tab:bleu2}
\end{table}

Tables~\ref{tab:bleu3} and \ref{tab:bleu2} show further experiments with the proposed methods on various English-German test sets, compared to several previous systems. Table~\ref{tab:bleu3} shows BLEU values calculated by \emph{multi-bleu}, and includes the NMT system proposed by \citet{D15-1166} which replaces unknown predicted words 
with the most strongly aligned word on the source sentence. 
Also, the table includes other systems described in Section~\ref{sec:related}. Additionally, Table~\ref{tab:bleu2} shows values calculated by the NIST BLEU scorer, as well as
results reported by the ``Winning WMT'' systems for each test set respectively: UEDIN-SYNTAX \cite{williams-EtAl:2014:W14-33}, UEDIN-SYNTAX \cite{williams-EtAl:2015:WMT}, and UEDIN-NMT \cite{sennrich-haddow-birch:2016:WMT}. Also, we include the results reported by \citet{sennrich-haddow-birch:2016:P16-12} for a baseline encoder-decoder NMT with BPE for unknown words similar to our configuration, and finally the system proposed by \citet{nadejde2017predicting}, an explicit syntax-aware NMT that introduces combinatory categorial grammar (CCG) supertags on the target side by predicting words and tags alternately. The comparison with this work is relevant for the analysis described later in Section~\ref{sec:structure}. 
The results confirm that the \emph{self-attentive residual connections} improve significantly the translations. To evaluate the significance of the improvements against the NMT baseline, we performed a one-tailed paired $t$-test.

\subsection{Impact of the Attention Function}
We now examine the two scoring functions that can be used for the \emph{self-attentive residual connections} model presented in Eq.~\eqref{eq:scoring-functions}, considering English-to-Chinese and Spanish-to-English. The BLEU scores are presented in Table~\ref{tab:score}: the best option is the \emph{content} matching function, which depends only on the word embeddings. The \emph{content+scope} function, which depends additionally on the hidden representation of the current prediction is better than the baseline but scores lower than \emph{content}. 
\begin{table}[t]
	\center
	\small
	{\def\arraystretch{1.1}\tabcolsep=3pt
		\begin{tabular}{l c c }
			 & \multicolumn{2}{c}{\textbf{BLEU}} \\ [1mm]  \hline
			\textbf{Attention function} & \textbf{En-Zh}& \textbf{Es-En} \\ [1mm]  \hline

		\emph{Content+Scope} & 23.1 & \text{25.6}    \\
		 \emph{Content} & \textbf{24.0} & \textbf{26.3}    \\ \hline
	\end{tabular}}
	\caption{BLEU scores for two scoring variants of the attention function of the proposed decoder.} 
	\label{tab:score}

\end{table}
The idea that the importance of the context depends on the current prediction is appealing, because it can be interpreted as learning internal dependencies among words. However, the experimental results show that it does not necessarily lead to the best translation. On the contrary, the \emph{content} attention function may be extracting representations of the whole sentence which are easier to learn and generalize.

\subsection{Performance According to Human Evaluation}

Manual evaluation on samples of 50 sentences for each language pair
helped to corroborate the conclusions obtained from the BLEU scores,
and to provide a qualitative understanding of the improvements
brought by our model. For each language, we employed
one evaluator who was a native speaker of the target
language and had good knowledge of the source language.
The evaluators ranked three translations of the same source
sentence -- one from each of our models: \emph{baseline}, \emph{mean
residual connections}, and \emph{self-attentive residual connections} --
according to their translation quality. The three translations
were presented in a random order, so that the system that had
generated them could not be identified. To integrate the judgments,
we proceed in pairs, and count the number of times
each system was ranked higher, equal to, or lower than another
competing system. The results shown in Table~\ref{tab:manual} indicate
that the \textit{self-attentive residual connections} model outperforms the one with \textit{mean
residual connections}, and both outperform the baseline, for
all three language pairs. The rankings are thus identical to
those obtained using BLEU in Tables~\ref{tab:bleu} and \ref{tab:bleu2}.

\begin{table}[t]
	\center
	\small
	{\def\arraystretch{1.1}\tabcolsep=2pt
		\begin{tabular}{l c c c | c c c | c c c}
			&   \multicolumn{9}{c}{\textbf{Ranking (\%)}} \\ \hline
			\textbf{System} & \multicolumn{3}{c|}{\textbf{En\textendash Zh}}& \multicolumn{3}{c}{ \textbf{Es\textendash En} }&  \multicolumn{3}{|c}{\textbf{En\textendash De}} \\ \hline
			& \textbf{$>$}& \textbf{$=$} & \textbf{$<$} & \textbf{$>$}& \textbf{$=$} & \textbf{$<$}& \textbf{$>$}& \textbf{$=$} & \textbf{$<$}\\  \hline
			Mean vs.\ Baseline  & 26 & 56 & 18 & 20 & 64 & 16 &28 & 58 & 24  \\
			Self-attentive vs.\ Baseline & 28 & 60 & 12 & 28 & 56 & 16 & 32 & 54 & 14  \\
			Self-attentive vs.\ Mean  & 24 & 62 & 14&28 &58 & 14 & 32 & 56 & 12  \\ \hline
	\end{tabular}}
	\caption{Human evaluation of sentence-level translation
		quality on three language pairs. We compare the models
		in pairs, indicating the percentages of sentences that were
		ranked higher ($>$), equal to ($=$), or lower ($<$) for the first
		system with respect to the second one. The values correspond to percentages (\%).}
	\label{tab:manual}
\end{table}

\begin{table}[t]
	\center
	\small
	{\def\arraystretch{1.1}\tabcolsep=1.2pt
		\begin{tabular}{l c c c c }
			\textbf{Systems} & $d$ &  \textbf{Perplexity} \\ [1mm]  \hline
		  	LSTM  \cite{daniluk2016frustratingly} & 300  & 85.2  \\
		  	LSTM + Attention \cite{daniluk2016frustratingly} & 296  & 82.0  \\
		  	LSTM + 4-gram  \cite{daniluk2016frustratingly} & 968 & 75.9  \\ \hdashline
			LSTM + Mean residual connections & 296 &  80.2  \\
			LSTM + Self-attentive residual connections   & 296  & 80.4  \\  \hline
	\end{tabular}}
	\caption{Evaluation of the proposed methods on language modeling. The number of parameter for all models is 47M.} 
	\label{tab:lm}

\end{table}

\subsection{Performance on Language Modeling}
To examine whether language modeling (LM) can benefit from the proposed method, we incorporate the residual connections into a neural LM. We use the same setting as \citet{daniluk2016frustratingly} for a corpus of Wikipedia articles (22.5M words), and we compare with two methods proposed in the same paper, namely attention LSTM and 4-gram LSTM. As shown in Table~\ref{tab:lm}, the proposed models outperform the LSTM baseline as well as the self-attention model, but not the 4-gram LSTM. Experiments using 4-gram LSTM for NMT showed poor performance (13.9 BLEU points for English-Chinese) which can be attributed to the difference between the LM and NMT tasks. Both tasks predict one word at a time conditioned over previous words, however, in NMT the previous target-word-inputs are not given, they have to be generated by the decoder. Thus, the output could be conditioned over previous erroneous predictions affecting in higher proportion the 4-gram LSTM model. This shows that even if a model improves language modeling, it does not necessarily improve machine translation.

\begin{figure}[t]
  \centering
  \includegraphics[width=0.9\linewidth]{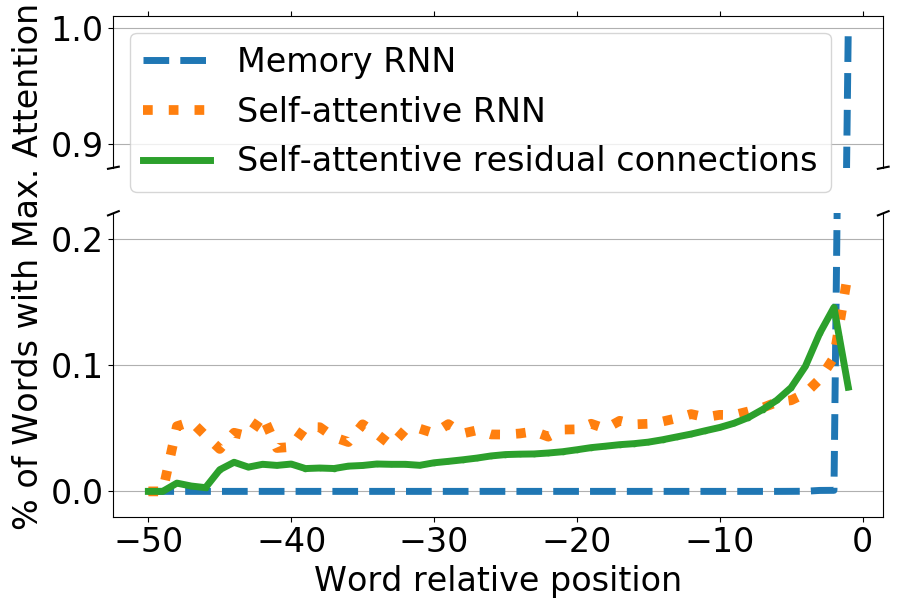}
 \caption{Percentage of words that received maximum attention at a given relative position, ranging from $-1$ to $-50$ (maximum length). 
 }
 \label{fig:context}
\end{figure}

\begin{figure*}[t!]
	\centering
	\begin{subfigure}[b]{0.32\textwidth}
		\includegraphics[width=1\textwidth, height=3.5cm]{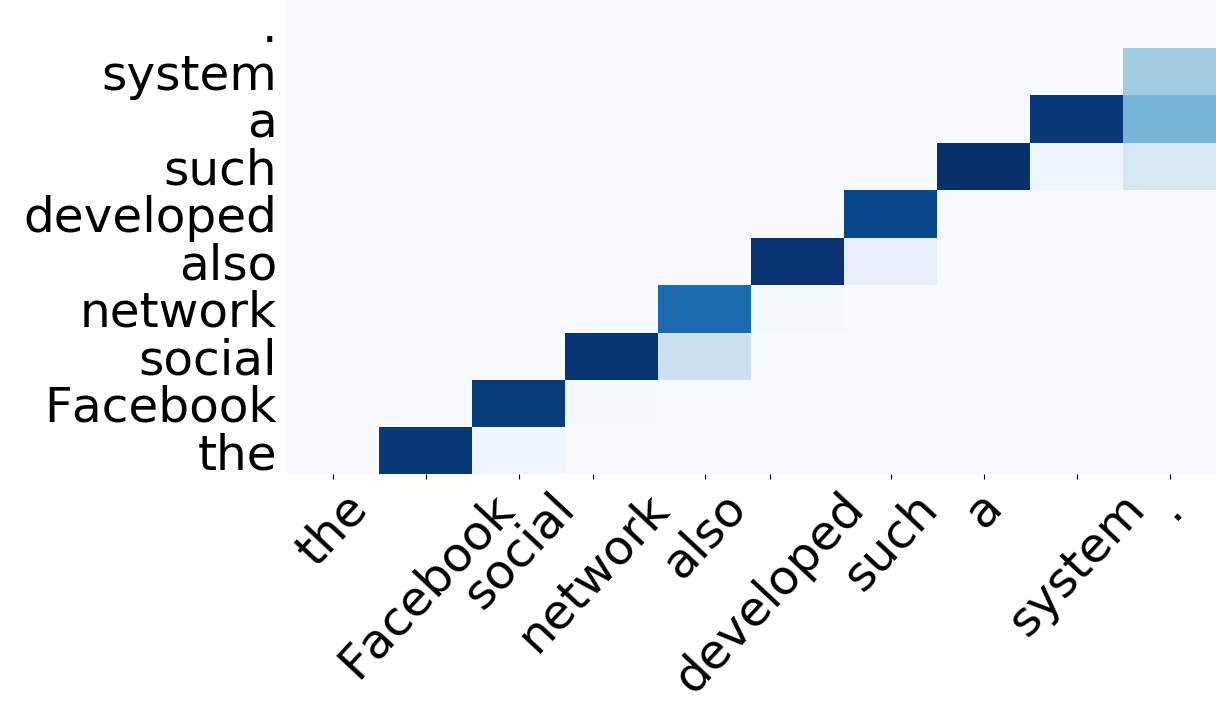}
		\caption{Memory RNN}
		\label{fig:m_att_mem}
	\end{subfigure}
	\begin{subfigure}[b]{0.32\textwidth}
		\includegraphics[width=1\textwidth, height=3.5cm]{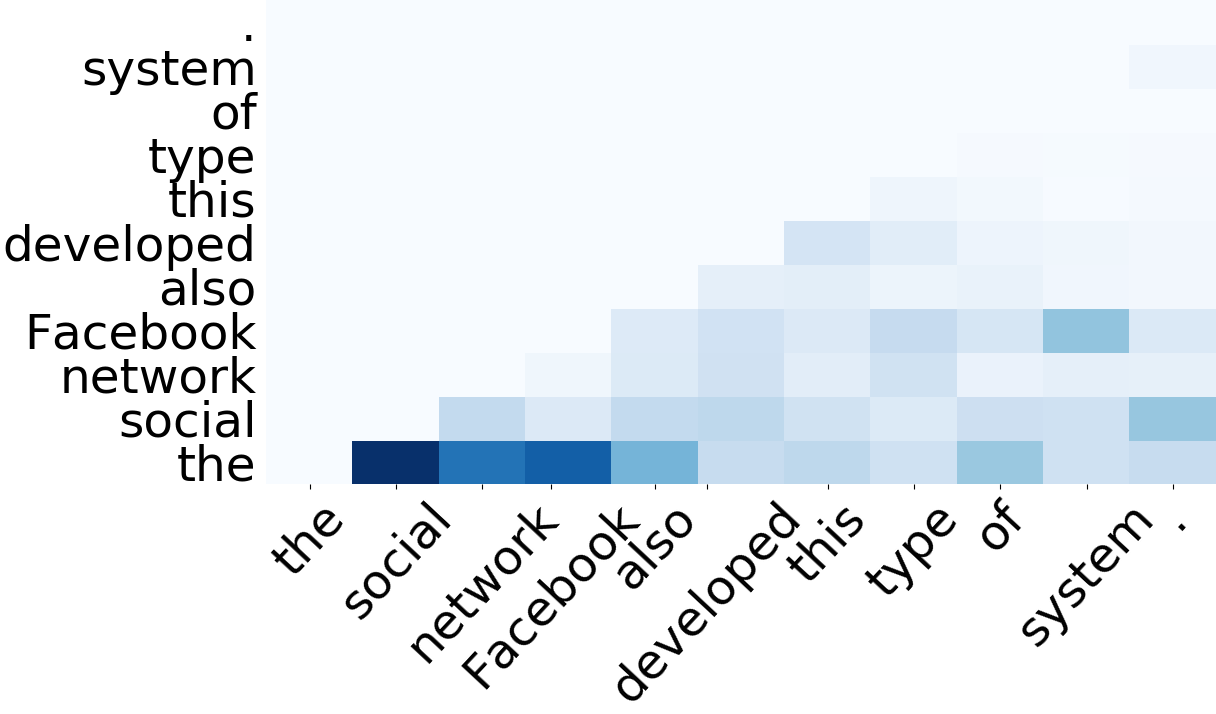}
		\caption{Self-attentive RNN}
		\label{fig:m_att_rep}
	\end{subfigure}
	\begin{subfigure}[b]{0.32\textwidth}
		\includegraphics[width=1\textwidth, height=3.5cm]{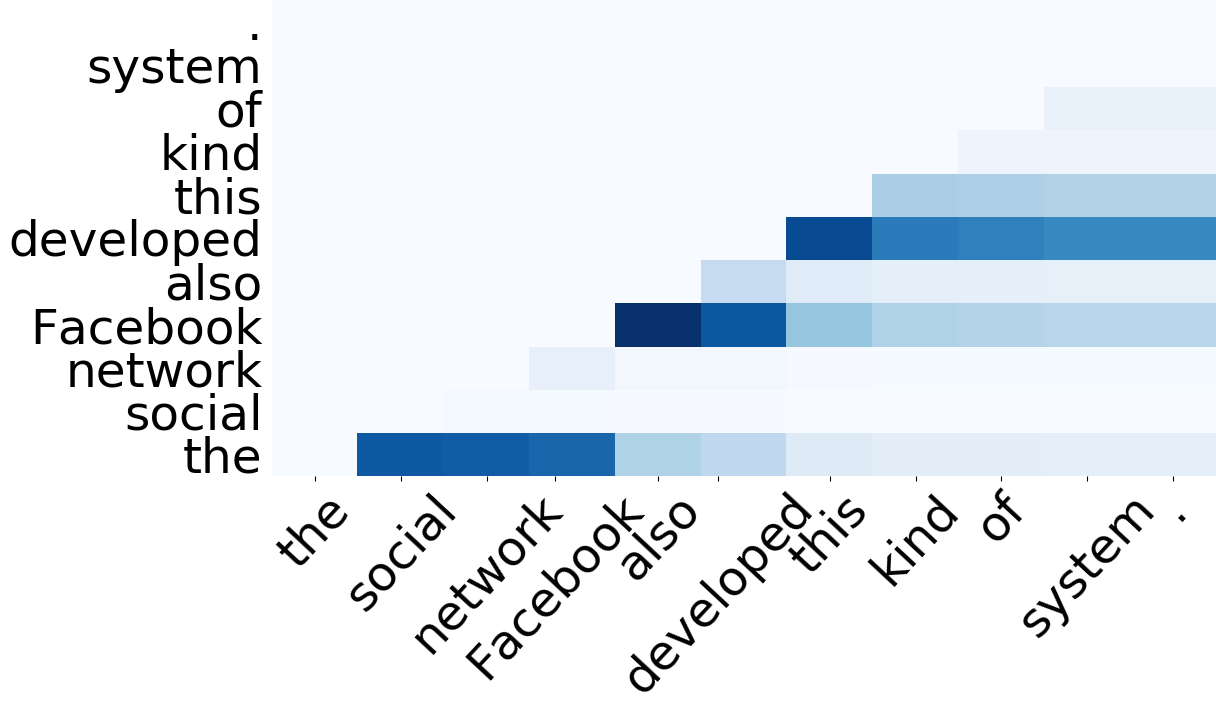}
		\caption{Self-attentive residual connections}
		\label{fig:m_att_res}
	\end{subfigure}
	\caption{Matrix of distribution of the attention weights to previous words. The vertical axis represents the previous words.  A darker shade indicates a higher attention weight.}
	\label{fig:sample}
	
\end{figure*}

\section{Qualitative Analysis}

\subsection{Distribution of Attention}
Figure~\ref{fig:context} shows a comparison of the distribution of attention of the different self-attentive models described in this paper,
on Spanish-to-English NMT (the other two language pairs exhibit similar distributions). The values correspond to the number of words which received maximal attention for each relative position ($x$-axis). We selected, at each prediction, the preceding word with maximal weight, and counted its relative position. We normalized the count by the number of previous words at the time of each prediction.

We observe that the \emph{memory RNN} almost always selects the immediately previous word ($t-1$) and ignores the rest of the context. On the contrary, the other two models distribute attention more evenly among all previous words. In particular, the \emph{self-attentive RNN} uses a longer context than the \emph{self-attentive residual connections} but, as the performance on BLEU score shows, this fact does not necessarily mean better translation.

Figure~\ref{fig:sample} shows the attention to previous words generated by each model for one sentence translated from Spanish to English. The matrices present the target-side attention weights, with the vertical axis indicating the previous words, and the color shades at each position (cell) representing  the attention weights. The weights of the \emph{memory RNN} 
are concentrated on the diagonal,
indicating that the attention is generally located on the previous word, which makes the model almost equivalent to the baseline. The weights of the \emph{self-attentive RNN} show that attention is more distributed towards the distant past, and they vary for each word because the attention function depends on the current prediction. This model tries to find dependencies among words, although complex relations seem difficult to learn. On the contrary, the proposed \emph{self-attentive residual connections} model strongly focuses on particular words, and we present a wider analysis of it in the following section.

\begin{algorithm}[t]
\small
\caption{Binary Parse Tree}\label{parse}
\begin{algorithmic}[1]
\Require $\textbf{A}$ matrix of attention of size $N \times N$
\Require $\textbf{s}$ sentence as list of words of size $N$
\Function{split}{$tree,\textbf{A},\textbf{s}$}
\State $n \gets length(s)$
\State $i \gets 0$
\While{$max(\textbf{A}[:][i])=0$ or $i<n$}
\State $i \gets i+1$
\EndWhile
\State $tree.addChild(\textbf{s}[0\colon i])$
\If{$i<n$}
\State $subtree  \gets new Tree()$
\State \Call{split}{$subtree, \textbf{A}[i\colon n][i\colon n], \textbf{s}[i\colon n]) $}
\State $tree.addChild(subtree)$
\EndIf
\EndFunction
\State $tree  \gets new Tree()$;  \Call{split}{$tree, \textbf{A}, \textbf{s} $}

\end{algorithmic}

\end{algorithm}

\begin{figure}[t]
	\centering
	\begin{tabular}{c}
	\includegraphics[width=0.98\linewidth]{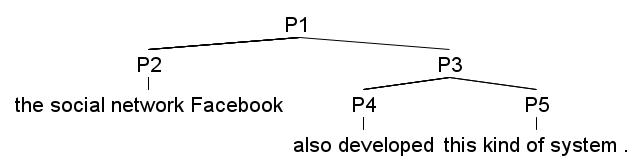}  \\ \\
	\includegraphics[width=0.98\linewidth]{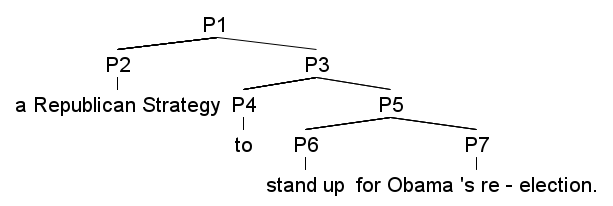}  \\ \\
	\includegraphics[width=0.98\linewidth]{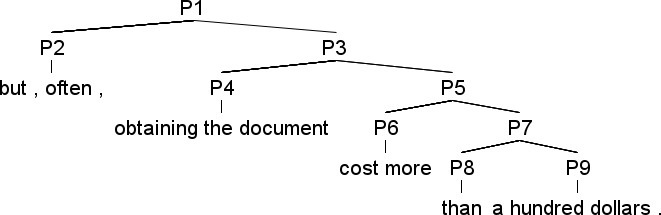} \\
	\end{tabular}
	\caption{Examples of hypothesized syntactic structures obtained with Algorithm~\ref{parse}.}
	\label{fig:tree}
\end{figure}

\subsection{Structures Learned by the Model}\label{sec:structure}
When visualizing the matrix of attention weights generated by our model (Figure~\ref{fig:m_att_res}), we observed the formation of sub-phrases which are grouped depending on their attention to previous words. 
To build the sub-phrases in a deterministic fashion, 
we implemented Algorithm~\ref{parse}, which iteratively splits the sentence into two sub-phrases every time the focus of attention changes to a new word, from left-to-right. The results are binary tree structures containing the sub-phrases, exemplified in Figure~\ref{fig:tree}.

We formally evaluate the syntactic properties of the binary tree structures by comparing them with the results of an automatic constituent parser \cite{manning-EtAl:2014:P14-5}, using the ParsEval approach \cite{H91-1060}, i.e.\ by counting the precision and recall of constituents, excluding single words. Our models reaches a precision of 0.56, which is better than the precision of 0.45 obtained by a trivial right-branched tree model\footnote{A model constructed by dividing iteratively one word and the rest of the sentence, from left-to-right.}. 
Note that these structures were neither optimized for parsing nor learned using part-of-speech tagging as most parsers do. Our interpretation of the results is that they are ``syntactic-like'' structures. However, given the simplicity of the model, 
they could also be viewed as more
limited structures, similar to sentence chunks.

\subsection{Translation Examples}
Table~\ref{tab:samples} shows examples of translations produced with the baseline and the \emph{self-attentive residual connections} model. The first part shows 
examples for which the proposed model reached a higher BLEU score than the baseline. Here, the structure of the sentences, or at least the word order, are improved. The second part contains 
examples where the baseline achieved better BLEU score than our model. In the first example, the structure of the sentence is different but the content and quality are similar, while in the second one lexical choices differ from the reference.

\begin{table}
	\centering
	\small
	{\def\arraystretch{1.2}\tabcolsep=2pt
		\begin{tabular}{|l p{0.9\linewidth}|}\hline
		\multicolumn{2}{|c|}{\textbf{Better than baseline}} \\ \hline
		S:& Estudiantes y profesores se est\'{a}n tomando a la ligera la fecha.\\
		R:& Students and teachers are taking the date lightly. \\
		B:& Students and teachers are \textcolor{red}{\textit{being taken lightly to the date}}. \\
		O:&Students and teachers are \textcolor{blue}{\textbf{taking the date lightly}}.  \\\hdashline

		S:& No porque compartiera su ideolog\'{\i}a, sino porque para \'{e}l los Derechos Humanos son indivisibles. \\
		R:& Not because he shared their world view, but because for him, human rights are indivisible.\\
		B:& Not because \textcolor{red}{\textit{I}} share his ideology, but because \textcolor{red}{\textit{he is indivisible by human rights}}.\\
		O:& Not because \textcolor{blue}{\textbf{he}} shared his ideology, but because \textcolor{blue}{\textbf{for him human rights are indivisible}}. \\\hline

		\multicolumn{2}{|c|}{\textbf{Worse than baseline}} \\ \hline
		S:& El gobierno intenta que no se construyan tantas casas peque\~{n}as. \\
		R:& The Government is trying not to build so many small houses. \\
		B:& The government is trying \textcolor{blue}{\textit{not to build so many small houses}}.\\
		O:& The government is trying \textcolor{red}{\textbf{to ensure that so many small houses are not built}}. \\\hdashline

		S:& Otras personas pueden tener ni\~{n}os . \\
		R:& Other people can have children. \\
		B:& \textcolor{blue}{\textit{Other people can}} have children.\\
		O:& \textcolor{blue}{\textbf{Others may}} have children. \\\hline

	\end{tabular}}
	\caption{Examples from Spanish to English.  } 
	\label{tab:samples}

\end{table}


\section{Conclusion}

We presented a novel decoder which uses self-attentive residual connections to previously translated words in order to enrich the target-side contextual information in  NMT.
To cope with the variable lengths of previous predictions, we proposed two methods for context summarization: \emph{mean residual connections} and \emph{self-attentive residual connections}. Additionally,
we showed how similar previous proposals, designed for language modeling, can be adapted to NMT.
We evaluated the methods over three language pairs: Chinese-to-English, Spanish-to-English, and English-to-German.  In each case, we improved the BLEU score compared to the NMT baseline and two variants with memory-augmented decoders. A manual evaluation over a small set of sentences for each language pair confirmed the improvement.
Finally, a qualitative analysis showed that the proposed model distributes weights throughout an entire sentence, and learns structures resembling syntactic ones.

As future work, we plan to enrich the present attention mechanism with the \emph{key-value-prediction} technique  \cite{daniluk2016frustratingly, miller2016key} which was shown to be useful for language modeling. Moreover, we will incorporate relative positional information to the attention function.
To encourage further research in \emph{self-attentive residual connections} for NMT an other similar tasks, our code is made publicly available\footnote{\url{https://github.com/idiap/Attentive_Residual_Connections_NMT}}. 

This work is part of the project \emph{Towards Document-Level Neural Machine Translation} \cite{miculicich2017towards}.

\section*{Acknowledgments}

We are grateful for support to the European Union under the Horizon 2020 SUMMA project (grant n.\ 688139, see \url{www.summa-project.eu}). We would also like to thank
James Henderson for his valuable feedback and suggestions. 

\bibliographystyle{acl_natbib_nourl}
\bibliography{Context_dec}

\end{document}